\documentclass[runningheads]{llncs}
\usepackage{graphicx}
\usepackage{amsmath}
\usepackage{float}
\usepackage[margin=1.5in]{geometry}    
\usepackage[english]{babel}
\usepackage[utf8]{inputenc}
\usepackage{algorithm}
\usepackage{arevmath}     
\usepackage[noend]{algpseudocode}
\begin{document}
\title{Design and Analysis of a 
Robotic Lizard using Five-Bar Mechanisms\thanks{Supported by Tentacles Robotic Foundation}}
%
%
\author{Rajashekhar V S \inst{1},
Dinakar Raj C K\inst{2}, Vishwesh S\inst{2}, Selva Perumal E\inst{2} and Nirmal Kumar M\inst{2}}
\authorrunning{Rajashekhar V S et al.}
%
\institute{Researcher, Tentacles Robotic Foundation,	Kanchipuram, Tamil Nadu, India, 603202 \and
Department of Mechanical Engineering, Adhiparasakthi Engineering College,	Melmaruvathur, Kanchipuram, Tamil Nadu, India, 603319\\
\email{vsrajashekhar@gmail.com}\\
}
\maketitle              
\begin{abstract}
Legged robots are being used to explore rough terrains as they are capable of traversing gaps and
obstacles. In this paper, a new mechanism is designed to replicate a robotic lizard using integrated five-
bar mechanisms. There are two five bar mechanisms from which two more are formed by connecting the
links in a particular order. The legs are attached to the links of the five bar mechanism such that, when
the mechanism is actuated, they move the robot forward. Position analysis using vector loop approach
has been done for the mechanism. A prototype has been built and controlled using servo motors to verify
the robotic lizard mechanism.
\end{abstract}
\section{Introduction}
Mimicking animals for the purpose of solving human centered problems plays an important role in the field of robotics. Among these four legged creatures are gaining attention in the past few decades \cite{siegwart2011introduction}. Among them MEMS based robots are gaining importance \cite{doshi2019contact}. Miniaturized robots can be used for surveillance and locomotion in cramped spaces.

A lizard is a commonly seen reptile whose mechanics have been studied in detail in the following
literature \cite{farley1997mechanics}. It was found that the walking and trotting gaits \cite{kim2013trotting} exhibited by the lizards were similar to that of the mammals. The biomechanics and kinematics of the sprawling pattern exhibited by the lizards were studied extensively in \cite{russell2001biomechanics}. The sprawling gait has be experimentally shown on a dynamic model in \cite{kim2014trot}. It is observed that the compliant and flexible trunk of the lizards help in reducing the peak power \cite{gu2015effects}. 

The lizard based bio-inspired robots have been made in the past, such as a water running robot \cite{floyd2006novel}, a leg mechanism \cite{dai2009biomimetics} and a quadrupeds \cite{park2008dynamic,park2009dynamic}. Motion analyis \cite{kim2012motion}, motion planning \cite{ratliff2009chomp} and gait planning \cite{son2010gait} based on kinematics of the Gecko have been done on robotic lizards. A seven degrees of freedom robotic Gecko has been developed for the experimental verification of the kinematic analysis \cite{nam2009kinematic}. 

\begin{figure}[H]
\begin{center}
\includegraphics[width=150pt]{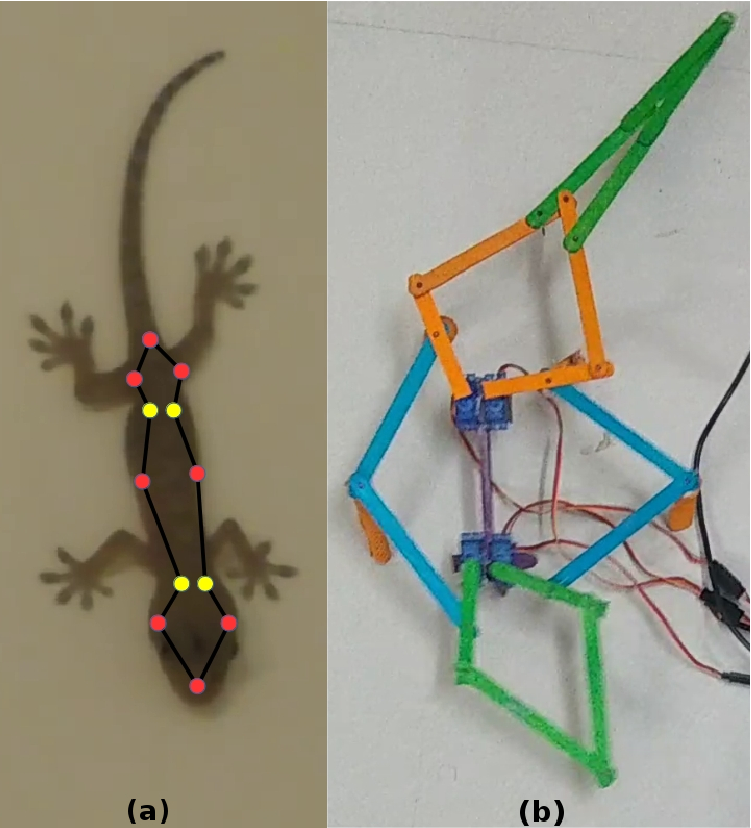}
\caption{(a) The real lizard that served as an inspiration (b) The bio-inspired robotic lizard introduced in this work} \label{fig_synthesis}
\end{center}
\end{figure}

Although there are works that exist in the literature that try to replicate a real lizard, they have not been able to mimic them closely. The novelty of this paper lies in the core mechanism design and it's abilty to replicate the gaits of a real lizard. The real lizard that served as a source of inspiration is shown in Figure \ref{fig_synthesis} (a) and the fabricated robotic lizard is shown in Figure \ref{fig_synthesis} (b). Topological design has been done using the method mentioned in \cite{yang2018topology}. The position analysis has been done using the vector loop approach. Finally the prototype of the robotic lizard has been shown exhibiting the walking gait.
\section{Topological Design of the Robotic Lizard Mechanism}
The design of robotic lizards using mechanisms like Watt-I Planar Linkage Mechanism \cite{xu2013bio} and CPG-driven modular robots \cite{vonasek2015high} have been done in the past. In this work, the development of the robotic lizard mechanism using two active and two passive five-bar mechanisms is done. The degree of freedom analysis is done to find out the number of degrees of freedom of the mechanism and also to find out whether the right driving links have been chosen for the mechanism. This is a eight step process and it is done using the method given in \cite{yang2018topology}.
\begin{figure}
\begin{center}
\includegraphics[width=350pt]{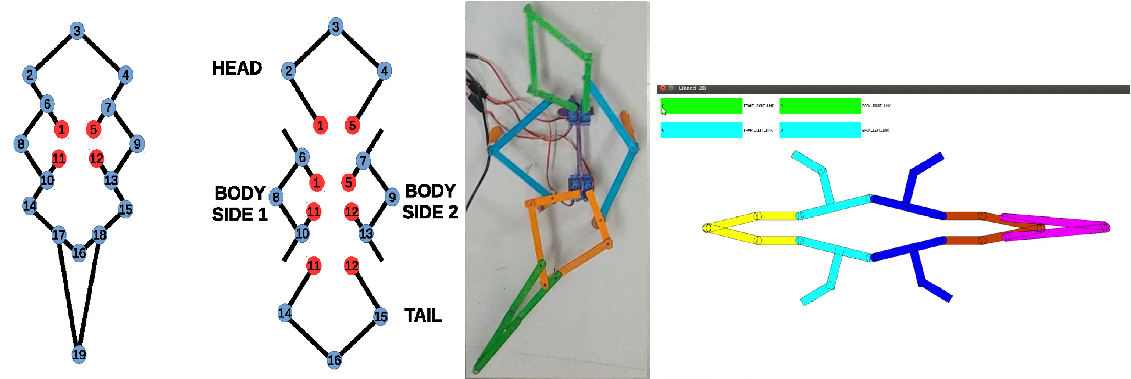}
\caption{(a) The topological design of the robotic lizard mechanism. (b) The four five-bar mechanisms that form the robotic lizard. It is used in position analysis (c) The prototype of the robotic lizard (d) The graphical user interface developed for the robotic lizard} \label{fig_topo}
\end{center}
\end{figure}
	\begin{equation}
\label{equ_maindof1}
F= \sum_{i=1}^{m} f_i - \sum_{j=1}^{v} \xi_{Lj}
\end{equation}
\begin{equation}
\label{equ_maindof2}
\sum_{j=1}^{v} \xi_{Lj} = dim.{((\cap_{i=1}^{j} M_{b_{i}})\cup M_{b_{j+1}})}
\end{equation}
where,\\
$F$ - Degree of freedom of parallel manipulator(PM).\\
$f_{i}$ - Degree of freedom of the $i^{th}$ joint.\\
$m$ - Total number of joints in the parallel manipulator.\\
$v$ - Total number of independent loops in the mechanism, where $v=m-n+1$.\\
$n$ - Total number of links in the mechanism.\\
$\xi_{Lj}$ - Total number of independent equations of the $j^{th}$ loop.\\
$\cap_{i=1}^{j} M_{b_{i}}$ -  Position and orientation characteristic (POC) set generated by the sub-PM formed by the former j branches.\\
$M_{b_{j+1}}$ - POC set generated by the end link of j+1 sub-chains.\\
Calculating the number of independent loops putting $n=13$ and $m=16$, we get $v=4$. The eight steps involved in the calculation of the degrees of freedom of the mechanism is as follows. 
\begin{enumerate}
\item The topological structure is mentioned symbolically here. \\
Branches:\\
$SOC_{1}({-R_{1}||R_{2}||R_{3}-})$; $SOC_{2}({-R_{4}||R_{5}-})$; $SOC_{3}({-R_{1}||R_{7}||R_{8}-})$;\\ $SOC_{4}({-R_{11}||R_{10}-})$; $SOC_{5}({-R_{5}||R_{7}||R_{9}-})$; $SOC_{6}({-R_{12}||R_{13}-})$;\\ $SOC_{7}({-R_{11}||R_{14}||R_{16}-})$; $SOC_{8}({-R_{12}||R_{15}-})$\\
Platforms:\\
Fixed platform: $R_{1},R_{5},R_{11},R_{12}$ \\
Moving platform: The rest are all moving platforms\\
\item An arbitrary point $o'$ is chosen on the moving platform. \\
\item 
Determining the POC set of branches\\
\[M_{b_{1}}=\begin{pmatrix} t^2 (\perp R_{i})\\
r^1 || (R_{i})\\
\end{pmatrix}\quad\text{i=1,2,3} \]
\[M_{b_{2}}=\begin{pmatrix} t^2 (\perp R_{j})\\
r^1 || (R_{j})\\
\end{pmatrix}\quad\text{j=4,5} \]
\item Finding the total number of independent displacement equations.\\  
Topological structure of the independent loop\\
$\xi_{L_{1}}$ = dim($M_{b_{1}}$ $\cup$ $M_{b_{2}})$= \[dim\{\begin{bmatrix}
t^2 \\
r^1 (|| R_{1})\\ 
\end{bmatrix}
\cup
\begin{bmatrix}
t^2 \\
r^1 (|| R_{4})\\   
\end{bmatrix}\} 
\]
 \[=dim\{\begin{bmatrix}
t^2 \\
r^1 (||\diamond (R_{1},R_{4}))\\ 
\end{bmatrix}\}\quad\text{=3}
\]
Similarly it can be done for the other 3 loops and the value of the number of independent displacement equations is found to be the same in each case.
\item Calculating the DOF of the mechanism\\
$F= \sum_{i=1}^{m} f_i - \sum_{j=1}^{v} \xi_{Lj} = 16 - (3+3+3+3) = 4$
\item Finding the inactive pairs. \\
Based on the calculations done in \cite{yang2018topology}, similar steps were followed. It was found that there is no inactive pair in the mechanism except for the tail which is regarded as passive linkages.
\item Determining the position and orientation characteristic set of the robotic lizard mechanism.\\
Based on the formula given in \cite{yang2018topology},\\
$M_{Pa}=M_{b_{1}} \cap M_{b_{2}}  $
\[M_{Pa}=\begin{bmatrix}
t^2 \\
r^1 (|| R_{1})\\ 
\end{bmatrix}
\cap
\begin{bmatrix}
t^2 \\
r^1 (|| R_{4})\\   
\end{bmatrix}
\cap
...
\cap
\begin{bmatrix}
t^2 \\
r^1 (|| R_{12})\\   
\end{bmatrix}
=
\begin{bmatrix}
t^2 \\
r^1 \\   
\end{bmatrix}
\]
The degree of freedom of the mechanism is 4 and the dimension of the above POC set is 3. Hence the module has two translational and one rotational degree of freedom. The discrepancy in the values is due to the fact that there is one more rotational degree of freedom on the same plane which the position and orientation characteristic analysis has ignored. This is because both the rotational degree of freedom are the same. 
\item Select driving pairs\\
The joints $R_{1}$, $R_{5}$, $R_{11}$  and $R_{12}$ are chosen to be the driving pairs and hence they are fixed. Calculating the degrees of freedom of the new mechanism, \\
$F^{*}= \sum_{i=1}^{m} f_i - \sum_{j=1}^{v} \xi_{Lj} = 12- (3+3+3+3) = 0$\\
Since $F^{*}=0$, the joints $R_{1}$, $R_{5}$, $R_{11}$ and $R_{12}$ can be used as driving pairs simultaneously. 
\end{enumerate}
Thus the topological design of the mechanism has been done. 
\section{Position Analysis of the Robotic Lizard Mechanism}
The robotic lizard mechanism is as shown in Figure \ref{fig_topo} (c). The vector loop method adopted from \cite{norton2004design} is used for the position analysis of the robotic lizard. There are five vector loop equations framed, one for the head, two for the body, one for the trunk and one for the tail. These equations are solved by keeping the known angles on one side and the unknown angles on the other side. The position of each part of the body is known by substituting the known four servo motor angles. These equations will be presented in detail in this section.\\
The vector loop equation of the head which forms a five-bar mechanism in the robotic lizard is written in Equation \ref{equ:head}. It is shown in Figure \ref{fig_topo} (b). 
\begin{equation}
\label{equ:head}
R_{2}+R_{3}-R_{4}-R_{5}-R_{1}=0
\end{equation}
Collecting the $\sin$ and $\cos$ terms together, the Equations \ref{equ:sin} and \ref{equ:cos} are obtained. 
\begin{equation}
\label{equ:sin}
L_{2}\sin(\theta_{2})+L_{3}\sin(\theta_{3})-L_{4}\sin(\theta_{4})-L_{5}\sin(\theta_{5})-L_{1}\sin(\theta_{1})=0
\end{equation}
\begin{equation}
\label{equ:cos}
L_{2}\cos(\theta_{2})+L_{3}\cos(\theta_{3})-L_{4}\cos(\theta_{4})-L_{5}\cos(\theta_{5})-L_{1}\cos(\theta_{1})=0 \\
\end{equation}
Solving for the $\theta_{3}$ and $\theta_{4}$ by rearranging the Equations \ref{equ:sin} and \ref{equ:cos}, the following are obtained. 
\begin{equation}
\label{equ:theta4}
\theta_4=2\arctan\frac{-B-\sqrt{B^{2}-4AC}}{2A}
\end{equation}
\begin{equation}
\label{equ:theta3}
\theta_3=2\arctan\frac{-E+\sqrt{E^{2}+4DF}}{2D}
\end{equation}
where,\\
$K_1=L_{5}\sin(\theta_{5})-L_{2}\sin(\theta_{2})$;\\
$K_2=L_{5}\cos(\theta_{5})-L_{2}\cos(\theta_{2})+L_{1}$\\
$K_3=-L_{5}\sin(\theta_{5})+L_{2}\sin(\theta_{2})$;\\
$K_4=-L_{5}\cos(\theta_{5})+L_{2}\cos(\theta_{2})+L_{1}$\\
$A=\frac{{L_{4}}^{2}-2\,K_2\,L4-{L_{3}}^{2}+{K_2}^{2}+{K_1}^{2}}{2}$;\\
$B=2\,K_1\,L_{4}$;\\
$C=\frac{{L_{4}}^{2}+2\,K_2\,L_{4}-{L_{3}}^{2}+{K_2}^{2}+{K_1}^{2}}{2}$\\
$D=\frac{{L_{4}}^{2}-{L_{3}}^{2}+2\,K_4\,L_{3}-{K_4}^{2}-{K_3}^{2}}{2}$;\\
$E=2\,K_3\,L_{3}$;\\
$F=\frac{{L_{4}}^{2}-{L_{3}}^{2}-2\,K_4\,L_{3}-{K_4}^{2}-{K_3}^{2}}{2}\\$
Similarly for the rest of the three five bar mechanisms, the vector loop equations are as follows. For the body side 1, body side 2 and tail,
\begin{equation}
R_{6}+R_{7}-R_{8}-R_{9}-R_{10}=0
\end{equation}
\begin{equation}
R_{12}+R_{13}-R_{14}-R_{15}-R_{11}=0
\end{equation}
\begin{equation}
R_{17}+R_{19}-R_{16}-R_{18}-R_{20}=0
\end{equation}
From the above equations, the values of $\theta$ can be found that are similar to Equations \ref{equ:theta4} and \ref{equ:theta3}. Thus the position analysis has been done for the mechanism. 
\subsection{Coordiates of Linkages in the Mechanism}
The four joints are actuated in a particular order so that the robotic lizard mechanism exhibits the forward walking gait. The equations governing the coordinates of the linkages of the head are as follows.
\begin{equation}
\begin{split}
&x_1=0\\
&y_1=0\\
&x_2=L_2 \cos(\theta_2)\\
&y_2=L_2 \sin(\theta_2)\\
&x_3=x_2+L_3 \cos(\theta_2+\theta_3)\\
&y_3=y_2+L_3 \sin(\theta_2+\theta_3)\\
&x_5=L_5 \cos(\theta_5)\\
&y_5=L_5 \sin(\theta_5)\\
&x_4=x_5+L_4 \cos(\theta_5+\theta_4)\\
&y_4=y_5+L_4 \sin(\theta_5+\theta_4)\\
\end{split}
\label{equ_fwdkin}
\end{equation}
In the similar way, it can be derived for the left and right side of the body, and the tail of the mechanism. 
\subsection{Workspace of robot parts}
\begin{figure}[H]
\begin{center}
\includegraphics[width=150pt]{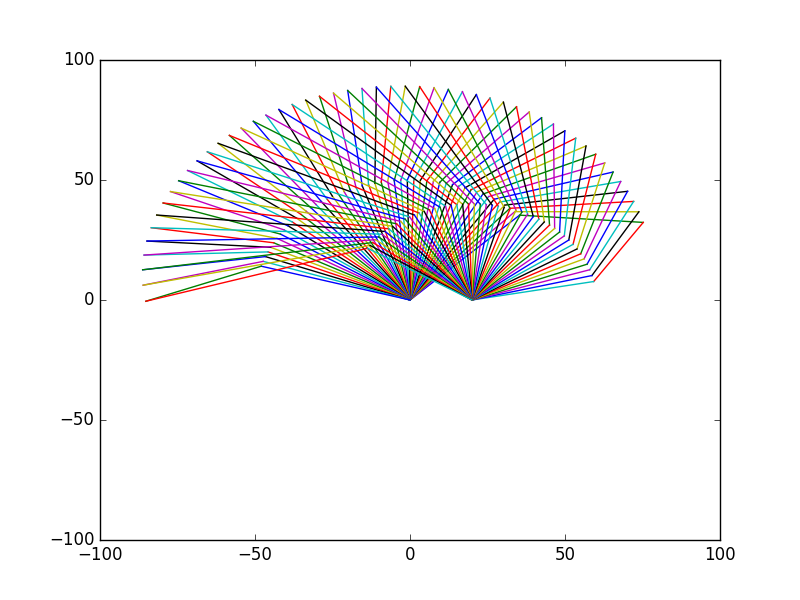}
(a)
\includegraphics[width=150pt]{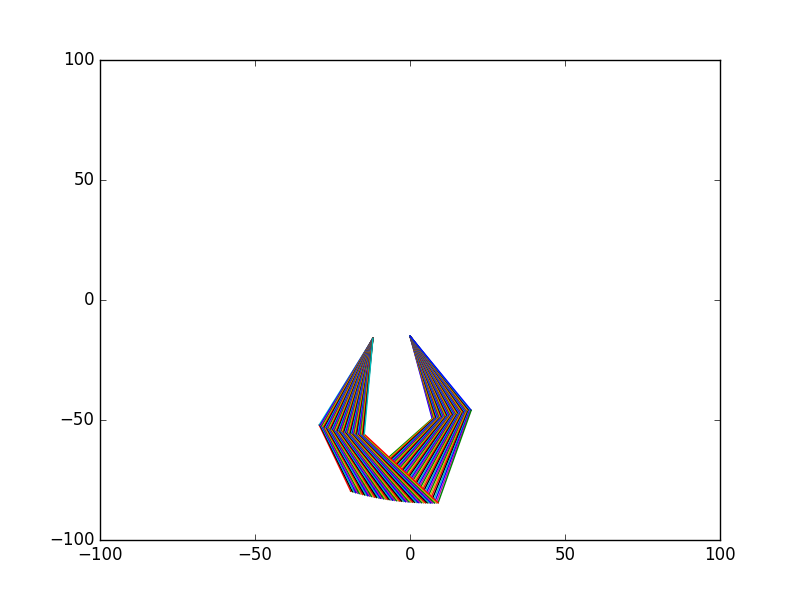}
(b)
\includegraphics[width=150pt]{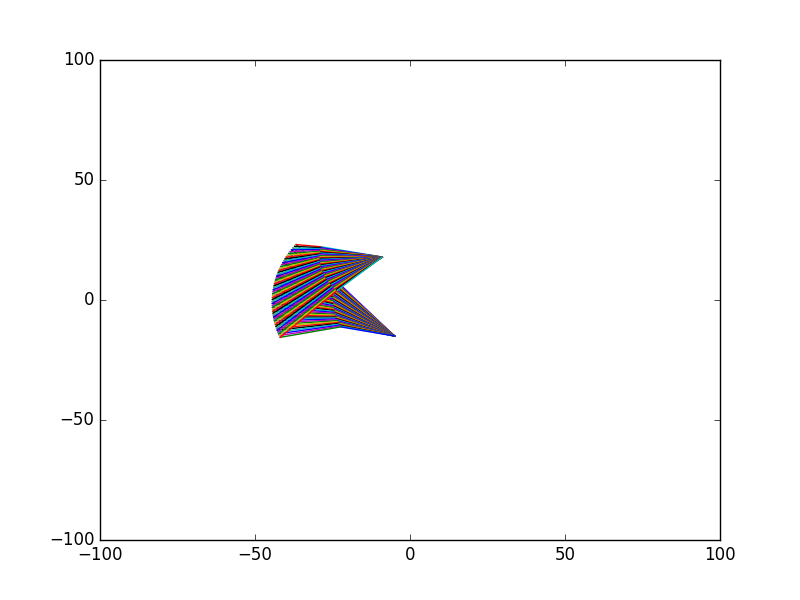}
(c)
\includegraphics[width=150pt]{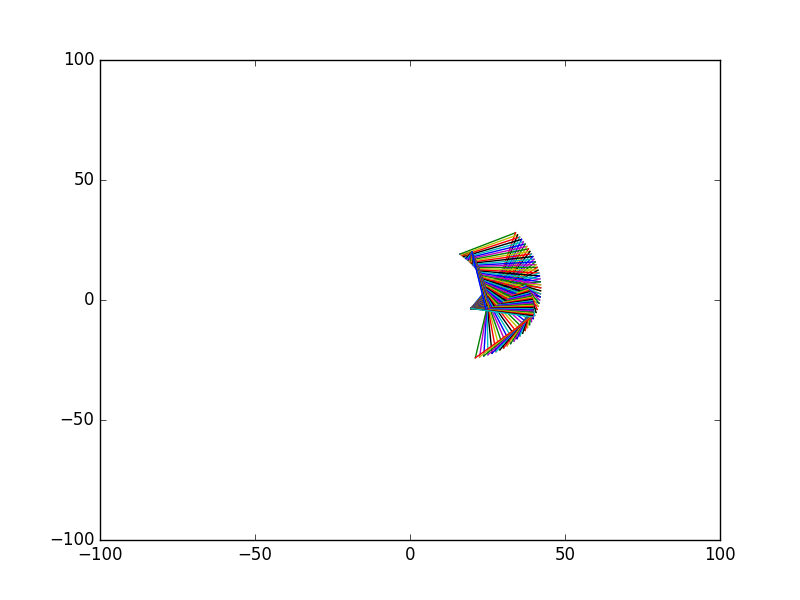}
(d)
\caption{The workspace of the (a) head (b) tail (c) body left (d) body right} 
\label{fig_workspace}
\end{center}
\end{figure}

The workspace of the robot is plotted by writing a $Python$ $Code$. The algorithm used for plotting the workspace of the robot is as follows. 
\begin{algorithm}[H]
\caption{Plotting the workspace using position analysis}
\begin{algorithmic}
\State Declare the range for $\theta_2$ and $\theta_5$ 
\State Declare the length of linkages
 \State For $\theta_5= 135$, $\theta_5{-}{-}$, while $\theta_5 < 0$
\State	  For $\theta_2= 45$, $\theta_2{+}{+}$, while $\theta_2 < 160$
\item 		Calculate Equation \ref{equ_fwdkin}
\item			plot([$x_1$,$x_2$],[$x_1$,$y_2$])\\
				plot([$x_2$,$x_3$],[$y_2$,$y_3$])\\
				plot([$x_3$,$x_4$],[$y_3$,$y_4$])\\
				plot([$x_4$,$x_5$],[$y_4$,$y_5$])
\item	 EndFor
\item EndFor
\end{algorithmic}
\end{algorithm}
The Figure \ref{fig_workspace} (a-d) shows the reachable workspace of the head, tail, body left and body right of the robotic lizard mechanism.   
\section{Prototype of the Robotic Lizard Mechanism}
\begin{figure}[H]
\begin{center}
\includegraphics[width=400pt]{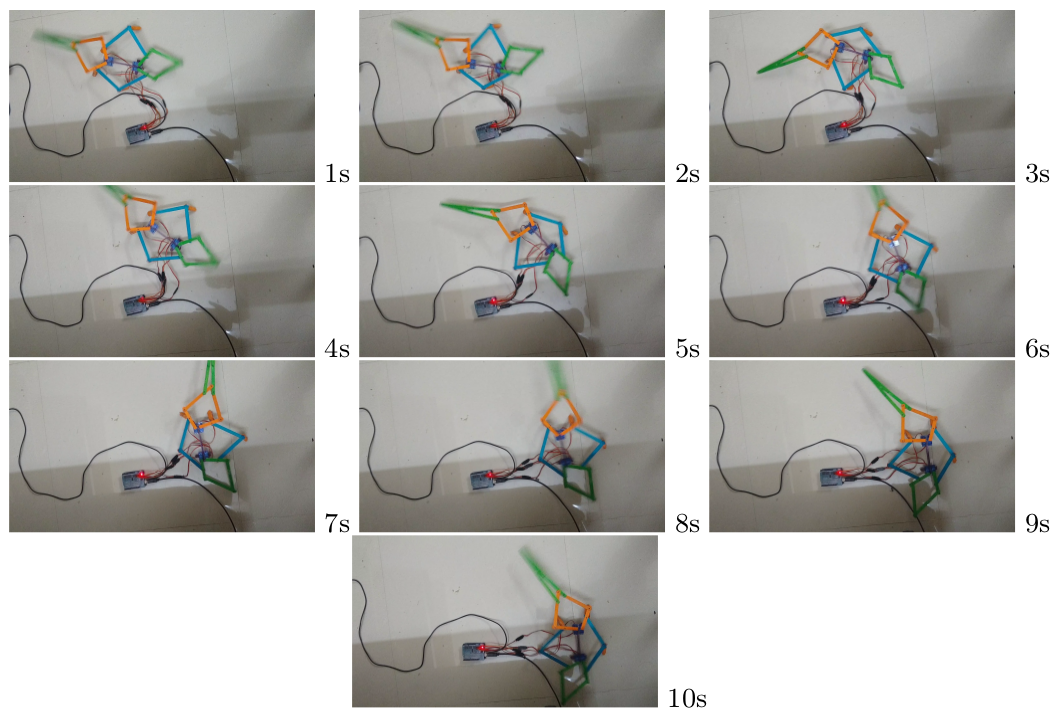}
\caption{The robotic lizard exhibiting the walking gait} \label{fig_tenfig}
\end{center}
\end{figure}

The links of the mechanism were fabricated using the balsa wood in a CNC router. The mechanism was
actuated using the four servo motors in a particular order. The controlled used here was Arduino UNO and
Processing IDE was used as a graphical user interface (GUI) to control the robot. The GUI is as shown in Figure \ref{fig_topo} (d). The robotic lizard was able to move on a flat surface and exhibit the walking gait as shown in Figure \ref{fig_tenfig}. 
\section{Conclusions}
In this work, a new robotic lizard mechanism was developed. The topological design of the mechanism was done. Then the position analysis of the mechanism was done to find the various angles between the links. Finally a prototype was made and controlled using servo motors. A graphical user interface was used to operate the robotic lizard. The future work would be to implement deep reinforcement learning to obtain a stable gait in unknown environments.  
\bibliographystyle{ieeetr}
\bibliography{mybibfile}

\begin{thebibliography}{10}

\bibitem{siegwart2011introduction}
R.~Siegwart, I.~R. Nourbakhsh, and D.~Scaramuzza, {\em Introduction to
  autonomous mobile robots}.
\newblock MIT press, 2011.

\bibitem{doshi2019contact}
N.~Doshi, K.~Jayaram, B.~Goldberg, Z.~Manchester, R.~J. Wood, and
  S.~Kuindersma, ``Contact-implicit optimization of locomotion trajectories for
  a quadrupedal microrobot,'' {\em arXiv preprint arXiv:1901.09065}, 2019.

\bibitem{farley1997mechanics}
C.~T. Farley and T.~C. Ko, ``Mechanics of locomotion in lizards.,'' {\em
  Journal of Experimental Biology}, vol.~200, no.~16, pp.~2177--2188, 1997.

\bibitem{kim2013trotting}
C.-H. Kim, H.-C. Shin, and H.-H. Lee, ``Trotting gait analysis of a lizard
  using motion capture,'' in {\em 2013 13th International Conference on
  Control, Automation and Systems (ICCAS 2013)}, pp.~1247--1251, IEEE, 2013.

\bibitem{russell2001biomechanics}
A.~Russell and V.~Bels, ``Biomechanics and kinematics of limb-based locomotion
  in lizards: review, synthesis and prospectus,'' {\em Comparative Biochemistry
  and Physiology Part A: Molecular \& Integrative Physiology}, vol.~131, no.~1,
  pp.~89--112, 2001.

\bibitem{kim2014trot}
C.~Kim, H.~Shin, and K.~Jeong, ``Trot gait simulation of four legged robot
  based on a sprawled gait,'' in {\em 2014 14th International Conference on
  Control, Automation and Systems (ICCAS 2014)}, pp.~1031--1036, IEEE, 2014.

\bibitem{gu2015effects}
X.~Gu, Z.~Guo, Y.~Peng, G.~Chen, and H.~Yu, ``Effects of compliant and flexible
  trunks on peak-power of a lizard-inspired robot,'' in {\em 2015 IEEE
  International Conference on Robotics and Biomimetics (ROBIO)}, pp.~493--498,
  IEEE, 2015.

\bibitem{floyd2006novel}
S.~Floyd, T.~Keegan, J.~Palmisano, and M.~Sitti, ``A novel water running robot
  inspired by basilisk lizards,'' in {\em 2006 IEEE/RSJ International
  Conference on Intelligent Robots and Systems}, pp.~5430--5436, IEEE, 2006.

\bibitem{dai2009biomimetics}
Z.~Dai, H.~Zhang, and H.~Li, ``Biomimetics of gecko locomotion: From biology to
  engineering,'' in {\em 2009 ASME/IFToMM International Conference on
  Reconfigurable Mechanisms and Robots}, pp.~464--468, IEEE, 2009.

\bibitem{park2008dynamic}
H.~S. Park, S.~Floyd, and M.~Sitti, ``Dynamic modeling of a basilisk lizard
  inspired quadruped robot running on water,'' in {\em 2008 IEEE/RSJ
  International Conference on Intelligent Robots and Systems}, pp.~3101--3107,
  IEEE, 2008.

\bibitem{park2009dynamic}
H.~S. Park, S.~Floyd, and M.~Sitti, ``Dynamic modeling and analysis of pitch
  motion of a basilisk lizard inspired quadruped robot running on water,'' in
  {\em 2009 IEEE International Conference on Robotics and Automation},
  pp.~2655--2660, IEEE, 2009.

\bibitem{kim2012motion}
C.-H. Kim, H.-C. Shin, and T.-w. Jeong, ``Motion analysis of lizard locomotion
  using motion capture,'' in {\em 2012 12th International Conference on
  Control, Automation and Systems}, pp.~2143--2147, IEEE, 2012.

\bibitem{ratliff2009chomp}
N.~Ratliff, M.~Zucker, J.~A. Bagnell, and S.~Srinivasa, ``Chomp: Gradient
  optimization techniques for efficient motion planning,'' 2009.

\bibitem{son2010gait}
D.~Son, D.~Jeon, W.~C. Nam, D.~Chang, T.~Seo, and J.~Kim, ``Gait planning based
  on kinematics for a quadruped gecko model with redundancy,'' {\em Robotics
  and Autonomous Systems}, vol.~58, no.~5, pp.~648--656, 2010.

\bibitem{nam2009kinematic}
W.~Nam, T.~Seo, B.~Kim, D.~Jeon, K.-J. Cho, and J.~Kim, ``Kinematic analysis
  and experimental verification on the locomotion of gecko,'' {\em Journal of
  Bionic Engineering}, vol.~6, no.~3, pp.~246--254, 2009.

\bibitem{yang2018topology}
T.-L. Yang, A.~Liu, H.~Shen, L.~Hang, Y.~Luo, and Q.~Jin, {\em Topology design
  of robot mechanisms}.
\newblock Springer, 2018.

\bibitem{xu2013bio}
L.~Xu, T.~Mei, X.~Wei, K.~Cao, and M.~Luo, ``A bio-inspired biped water running
  robot incorporating the watt-i planar linkage mechanism,'' {\em Journal of
  Bionic Engineering}, vol.~10, no.~4, pp.~415--422, 2013.

\bibitem{vonasek2015high}
V.~Von{\'a}sek, M.~Saska, L.~Winkler, and L.~P{\v{r}}eu{\v{c}}il, ``High-level
  motion planning for cpg-driven modular robots,'' {\em Robotics and Autonomous
  Systems}, vol.~68, pp.~116--128, 2015.

\bibitem{norton2004design}
R.~L. Norton {\em et~al.}, {\em Design of machinery: an introduction to the
  synthesis and analysis of mechanisms and machines}.
\newblock Boston: McGraw-Hill Higher Education,, 2004.

\end{thebibliography}
\end{document}